\documentclass{article}
\usepackage{amsmath}
\usepackage{amsfonts}
\usepackage{graphicx}
\usepackage{enumitem}

\usepackage[style=numeric]{biblatex}
\usepackage{xcolor}
\usepackage{hyperref}
\hypersetup{colorlinks=true,
linkcolor = blue,
linkbordercolor = {white},
citecolor = blue}

\DeclareMathOperator*{\argmax}{arg\,max}

\usepackage{mathtools}

\DeclarePairedDelimiterX{\infdivx}[2]{(}{)}{%
  #1\;\delimsize\|\;#2%
}

     \usepackage[nonatbib, final]{neurips_2020}

\usepackage[utf8]{inputenc} 
\usepackage[T1]{fontenc}    
\usepackage{hyperref}       
\usepackage{url}            
\usepackage{booktabs}       
\usepackage{amsfonts}       
\usepackage{nicefrac}       
\usepackage{microtype}      

\title{Lagrangian Duality in Reinforcement Learning}

\author{%
  Pranay Pasula \\
  Department of Electrical Engineering and Computer Sciences\\
  University of California, Berkeley\\
  \texttt{pasula@berkeley.edu} \\
}

\begin{document}

\maketitle

\begin{abstract}
  Although duality is used extensively in certain fields, such as supervised learning in machine learning, it has been much less explored in others, such as reinforcement learning (RL). In this paper, we show how duality is involved in a variety of RL work, from that which spearheaded the field, such as Richard Bellman's value iteration, to that which was done within just the past few years yet has already had significant impact, such as TRPO, A3C, and GAIL. We show that duality is \emph{not} uncommon in reinforcement learning, especially when \emph{value iteration}, or \emph{dynamic programming}, is used or when first or second order approximations are made to transform initially intractable problems into tractable convex programs.
\end{abstract}

\section{Introduction}
The study of optimization problems that are dual to certain initial problems has led to key insights, including efficient ways to bound or exactly solve these original problems \cite{boyd2004convex}. Though duality is used extensively in certain fields, such as supervised learning in machine learning, it has been less explored in others, such as reinforcement learning (RL). In this paper, we show how duality is involved in a variety of RL work, from that which spearheaded the field \cite{bellman1966dynamic, howard1960dynamic, bertsekas1995dynamic} to that which was done within the past few years but has already had significant impact \cite{schulman2015trust, ho2016generative, mnih2016asynchronous}.

We show that duality is \emph{not} uncommon in reinforcement learning, especially when \emph{value iteration}, or \emph{dynamic programming}, is used or when first or second order approximations are made to transform initially intractable problems into tractable convex programs \cite{schulman2015trust, kakade2002natural, achiam2017constrained}. We touch on a number of works that span a wide range of RL paradigms but focus on some works that have been particularly influential, such as Trust Region Policy Optimization (TRPO) \cite{schulman2015trust}, Asynchronous Advantage Actor-Critic (A3C) \cite{mnih2016asynchronous}, and Generative Adversarial Imitation Learning (GAIL) \cite{ho2016generative}. In some cases duality is used as a theoretical tool to prove certain results or to gain insight into the meaning of the problem involved. In other cases duality is leveraged to employ gradient-based methods over some \emph{dual} space, as is done in alternating direction method of multipliers (ADMM) \cite{boyd2011distributed}, mirror descent \cite{beck2003mirror}, and dual averaging \cite{xiao2010dual, duchi2011dual}.

We hope this work will encourage others to more strongly consider duality in reinforcement learning, which we believe to be an often underexplored yet promising line of reasoning.

\section{Preliminaries}

We consider a standard reinforcement learning framework in which we have a Markov decision process (MDP) $M = (\mathcal{S}, \mathcal{A}, P, R, T, \gamma)$, where $\mathcal{S}$ is the state space, $\mathcal{A}$ is the action space, $P:\mathcal{S} \times \mathcal{A} \times \mathcal{S} \rightarrow [0, 1]$ is the distribution representing the transition probabilities of the agent arriving in state $s'$ after taking action $a$ while in state $s$, $R:\mathcal{S} \times \mathcal{A} \rightarrow \mathbb{R}$ is the reward function representing the reward  the agent obtains by taking action $a$ while in state $s$, $T > 0$ is the time horizon and can be either finite or infinite, and $\gamma \in [0,1]$ is the multiplicative rate at which future rewards are discounted per time step. For simplicity, we will usually set $\gamma = 1$ and drop this term. Throughout this paper we will use the terms \emph{trajectory}, \emph{behavior}, and \emph{demonstration} interchangeably, all of which refer to a finite or infinite sequence of states and actions $\{s_1, a_1, s_2, a_2, \dotsc, s_T, a_T\}.$

We assume the reader has at least an introductory understanding of duality theory and defer its exposition to \cite{boyd2004convex}.

\section{Duality in Dynamic Programming}

\subsection{Unregularized Markov Decision Processes}

Duality has well-established history in the theory underlying fundamental classes of RL algorithms, such as value iteration and policy optimization. In particular, the Bellman backup operator $\mathcal{T}_\pi$ \cite{bellman1966dynamic} induced by a policy $\pi$ and applied to Q functions $Q_\pi$ induced by $\pi$ represents the dynamic programming update
$$Q_{\pi}(s_t, a_t) = R(s_t, a_t) + \sum_{s_{t+1} \in \mathcal{S}} \left[ P(s_{t+1} \mid s_t, a_t) \sum_{a_{t+1} \in \mathcal{A}} Q_{\pi}(s_{t+1}, a_{t+1}) \right],$$
which under mild conditions \cite{puterman2014markov} converges to the fixed point $Q_{\pi}.$

Since the goal in reinforcement learning is to find the optimal policy $\pi^*$ that maximizes the expected sum of rewards obtained by the agent, an equivalent problem is to find $\pi^*$ that maximizes the average reward $\bar{R}$ obtained by the agent over timesteps $t = 1, 2, 3, \dotsc$

MDP theory allows us to write the average reward $\bar{R}(\pi)$, obtained by following policy $\pi$, in terms of $R$, $\pi$, and the stationary state distribution $\nu_\pi(x)$ under $\pi$. Namely, we have
$$\bar{R}(\pi) = \sum_{(s,a) \in \mathcal{S} \times \mathcal{A}} \nu_\pi(s)\pi(a \mid s) R(s,a).$$
The stationary state distribution $\nu_\pi$ is linear in the stationary state-action distribution $\mu_\pi = \nu_\pi \pi$, which suggests that the problem of finding the optimal policy can be formulated as a linear program (LP) with decision variable $\mu$ over the probability simplex $\mathcal{C}$:
\begin{equation}\label{sa_dist_opt}
\mu^* = \argmax_{\mu \in \mathcal{C}} \bar{R}(\mu).
\end{equation}
\cite{puterman2014markov} shows that by assuming $\mu^* \in \mathcal{C}$, which holds when the MDP has only one recurrent class, (\ref{sa_dist_opt}) is the dual of the LP
\begin{equation}\label{reward_opt}
\begin{split}
    & \underset{\bar{R} \in \mathbb{R}}{\max} \quad \qquad \bar{R}  \\
    & \text{subj. to } \quad \bar{R} - R(s,a) + V(s) - \sum_{s'} P(s' \mid s,a) V(s'), \qquad \text{for all } s,a \in \mathcal{S} \times \mathcal{A},
\end{split}
\end{equation}
where the dual variable $V(\cdot) = \sum_{a} P(a \mid \cdot{}) Q({}\cdot{}, a)$ is known as the \emph{value function}. Since this is a linear program, if the optimal average reward $\bar{R}^*$ is attained, strong duality holds, and it can be shown that the optimal value function $V^*$ is the solution to the \emph{average-reward Bellman equations}
\begin{equation}\label{avg_bell}
V^*(s) = \underset{a}{\max} \left( R(s,a) - \bar{R}^* + \sum_{s'} P(s' \mid s,a) V^*(s) \right), \quad \text{for all } s \in \mathcal{S},
\end{equation}
and that $V^*$ is a fixed point of the Bellman operator $\mathcal{T}_{\pi^*}$ under $\pi^*.$

\subsection{Regularized Markov Decision Processes}

\cite{neu2017unified} defines a regularized objective $\bar{R}_\eta(\mu)$ inspired by the linear program (\ref{reward_opt}) and proposes the concave optimization problem
\begin{equation}\label{reg_opt}
\underset{\mu \in \mathcal{C}}{\max} \ \bar{R}_\eta(\mu) = \underset{\mu \in \mathcal{C}}{\max} \ \left\{ \sum_{s,a} \big[ \mu(s,a)R(s,a) \big] - \frac{1}{\eta}W(\mu) \right\},
\end{equation}
where $\eta > 0$ is the scale parameter that trades-off between the original objective and regularization and $W : \mathbb{R}^{\mathcal{S} \times \mathcal{A}} \rightarrow \mathbb{R}$ is a convex regularization function.

\cite{neu2017unified} shows that by using Bregman divergences $D_S\infdivx{\mu} {\mu'}$, induced by the \emph{negative Shannon entropy}, and $D_C\infdivx{\mu}{ \mu'}$, induced by the \emph{conditional entropy}, for some reference distribution $\mu'$ as choices of $W(\mu)$ in (\ref{reg_opt}), the dual to (\ref{reg_opt}) is similar to the average-reward Bellman equations (\ref{avg_bell}). To note, these choices of $W$ allow us to perform \emph{mirror descent} to find both the primal and dual optimal values.

Furthermore, \cite{neu2017unified} shows that the popular reinforcement learning algorithms Trust Region Policy Optimization (TRPO) \cite{schulman2015trust} and Asynchronous Advantage Actor-Critic (A3C) \cite{mnih2016asynchronous} can be formulated as optimization problems with the form of (\ref{reg_opt}) for particular choices of $W$. We will revisit each of these when we discuss TRPO in Section 5.1 and A3C in Section 5.2.

\subsubsection{Relative entropy Bregman divergence}

Namely, \cite{peters2010relative, zimin2013online, neu2017unified} show that for $W := D_S\infdivx{\cdot}{\mu'}$, we have 
\begin{equation}\label{rel_ent_opt_dist}
\mu_\eta(s,a)^* \propto \mu'(s,a)e^{\eta \big(R(s,a) + \sum_{s'} P(s' \mid s,a) V^*(s') - V^*(s) \big)},
\end{equation}
and the dual function
\begin{equation}\label{rel_ent_dual_func}
g(V) = \frac{1}{\eta} \log \sum_{s,a} \mu'(s,a)e^{\eta \big(R(s,a) + \sum_{s'} P(s' \mid s,a) V^*(s') - V^*(s) \big)}.
\end{equation}
Assuming that strong duality holds, we have
\begin{equation}\label{strong_duality_rel_ent}
g(V^*_\eta) = \bar{R}^*_\eta = \max_{\mu \in \mathcal{C}} \bar{R}_\eta(\mu)
\end{equation}

\subsubsection{Conditional entropy Bregman divergence}

\cite{neu2017unified} shows that for $W := D_C\infdivx{\cdot}{\mu'}$, we have
\begin{equation}\label{cond_ent_opt_pol}
\pi_\eta^*(a \mid s) \propto \pi_{\mu'}(a \mid s)e^{\eta \big(R(s,a) + \sum_{s'} P(s' \mid s,a) V^*(s') - V^*(s) \big)}
\end{equation}
and the dual problem of (\ref{reg_opt}) is
\begin{equation}\label{dual_cond_ent_prob}
\begin{split}
    \underset{\lambda \in \mathbb{R}}{\max} \  & \quad \lambda \\
    \text{subj. to } & \quad V(s) = \frac{1}{\eta} \log{\sum_{a} \pi_{\mu'}(a \mid s)e^{\eta \big( R(s,a) - \lambda + \sum_{s'} P(s' \mid s,a)V(s') \big)}}, \quad \text{for all } s \in \mathcal{S},
\end{split}
\end{equation}
which has dual optimal solution
\begin{equation}\label{dual_cond_end_soln}
V^*_\eta (s) = \frac{1}{\eta} \log{\sum_{a} \pi_{\mu'}(a \mid s)e^{\eta \big( R(s,a) - \bar{R}^*_\eta + \sum_{s'} P(s' \mid s,a) V^*_\eta(s') \big)}}, \quad \text{for all } s' \in \mathcal{S}.
\end{equation}

\section{Duality in Learning from Demonstration}

Learning from demonstration (LfD) is a paradigm in which an agent is given demonstrations from some presumably expert agent and aims to imitate those demonstrations as closely as possible or learn the reward function that the expert was trying to maximize. LfD can be broken into three subfields, behavioral cloning, adversarial imitation learning, and inverse reinforcement learning. 

In behavioral cloning (BC), the agent aims to mimic the expert demonstrations without reward signal nor interaction with the environment. Though attractively simple, BC is susceptible to compounding errors due to covariate shift and lack of feedback. Since an agent that performs BC learns only from the demonstrations provided to it, BC usually requires a large amount of data that captures both the behaviors under optimal conditions as well as corrective actions that rectify suboptimal behaviors.

In adversarial imitation learning, a discriminator is given unlabeled demonstrations from the agent and expert. For each demonstration the discriminator is trained to classify whether the demonstration was generated by the agent or by the expert. The agent's goal is to fool the discriminator as often as possible, and by training to maximize this objective, the agent learns to generate demonstrations that are similar to the expert demonstrations. This is akin to the generative adversarial network (GAN) framework \cite{goodfellow2014generative}, and indeed, we will soon see that adversarial imitation learning can be see as GAN training \cite{ho2016generative}. Unlike in behavioral cloning, in adversarial imitation learning the agent generates trajectories by interacting with the environment.

In inverse reinforcement learning (IRL), the primary objective is to recover the reward function that the expert agent was optimizing for. Often simultaneously, an agent is trained to maximize an iteration of this reward function and generates demonstrations comparable to expert demonstrations at convergence. A major downside to IRL is that it is typically posed as a two-loop problem, in which the inner loop requires performing an expensive reinforcement learning procedure.

\subsection{Generative Adversarial Imitation Learning (GAIL)}

Generative Adversarial Imitation Learning \cite{ho2016generative} is a framework for learning an expert policy from expert demonstrations while avoiding the need to recover the optimal reward function. However, it assumes that the expert agent has optimized for some optimal reward function $R^*:\mathcal{S} \times \mathcal{A} \longrightarrow \mathbb{R}$ from which a unique optimal policy $\pi^*:\mathcal{S} \longrightarrow \mathcal{A}$ can be learned. In fact, since we assume that $R^* \in \text{IRL}(\pi^*)$ and $\pi^* \in \text{RL}(R^*)$, we have
$$R^* \in \argmax_{R \in \mathcal{R}} \ \mathbb{E}_{\pi^*} \left[ \sum_{t=0}^\infty \gamma^t R(s_t, a_t) \right].$$
$$\pi^* \in \argmax_{\pi \in \Pi} \ \mathbb{E}_{\pi} \left[ \sum_{t=0}^\infty \gamma^t R^*(s_t, a_t) \right].$$
IRL optimizes for $R^*$ and $\pi^*$ simultaneously by solving the saddle-point optimization problem
$$ \underset{R \in \mathcal{R}}{\max} \left( \underset{\pi \in \Pi}{\min} \  \mathbb{E}_{(s,a) \sim \pi} \left[ R(s,a) \right] \right) - \mathbb{E}_{(s,a) \sim \pi^*} \left[ R(s,a) \right]. $$
Maximum causal entropy IRL \cite{ziebart2008maximum, ziebart2010modeling} aims to recover a policy that not only performs well but also captures diverse behaviors. It does so by modifying the objective directly above to also maximize the causal entropy $H$ of the recovered policy.

\cite{ho2016generative} shows that under mild conditions, the dual to this modified problem is actually a problem that aims to match the stationary state-action distributions of the expert policy and the policy being trained. Through this result, they propose an adversarial imitation learning framework with guaranteed convergence results as well as a robust, practical imitation learning algorithm.

\section{Duality in Policy Search}

\subsection{Trust Region Policy Optimization (TRPO)}

Policy gradient (PG) algorithms \cite{sutton2000policy, silver2014deterministic} are a family of algorithms that aim to recover the optimal policy $\pi^*$ with respect to some reward function $R$, usually by interacting with the environment, obtaining Monte Carlo estimates of the discounted rewards, and computing the gradient of the policy with respect to these rewards so that gradient-based optimization over the policy space converges to $\pi^*$.

In vanilla PG, the policy being trained undergoes gradient updates that are based on a loss function with respect to the policy parameters. Though the policy is kept close in parameter space, in practice vanilla PG can be highly unstable, and policy performance can vary significantly from iteration to iteration. \cite{schulman2015trust} proposes Trust Region Policy Optimization (TRPO), which overcomes this issue by constraining the expected Kullback-Leibler divergence from the old policy $\theta^{k+1}$ to the new policy $\theta^{k}$ over the states encounted while running the old policy. Namely, \cite{schulman2015trust} proposes a heuristic approximation to the solution of the theoretical TRPO problem,
\begin{equation}\label{trpo_objec_1}
\begin{split}
\theta_{k+1} =  \quad \argmax_{\theta \in \Theta} \quad & \mathcal{L}(\theta, \theta^{k}) \\
\quad \text{subj. to } \quad & \bar{D}_{KL}\infdivx{\theta}{\theta^{k}} \leq \delta,
\end{split}
\end{equation}
where the \emph{surrogate advantage} function $\mathcal{L}(\theta, {\theta^k})$ quantifies how well the policy $\pi_\theta$ performs compared with the old policy $\pi_{\theta^k}$ using trajectories obtained by running the old policy,
$$\mathcal{L}(\theta, {\theta^k}) = \underset{(s,a) \sim \pi_{{\theta^k}}}{\mathbb{E}} \left[ \frac{\pi_\theta(a \mid s)}{\pi_{\theta^k}(a \mid s)} \left( R(s,a) - \underset{a \sim \pi_{\theta^k}(\cdot \mid s)}{\mathbb{E}} \left[ R(\cdot \mid s) \right] \right) \right],$$
and 
$$\bar{D}_{KL}\infdivx{\theta}{\theta^{k}} = \underset{s \sim \pi_{\theta^k}}{\mathbb{E}} \left[ D_{KL}\infdivx*{\pi_\theta(\cdot \mid s)}{\pi_{\theta^k}(\cdot \mid s)} \right].$$
Interestingly, by using a linear approximation to the objective function and a quadratic approximation to the constraint in (\ref{trpo_objec_1}), we get exactly the natural policy gradient \cite{kakade2002natural} update,
\begin{equation}\label{nat_grad}
\begin{split}
\theta^{k+1} = \quad \argmax_{\theta \in \Theta} \quad & \nabla_\theta \mathcal{L}(\theta, \theta^k) \Big\rvert_{\theta = \theta^k} \cdot (\theta - \theta^k) \\
\quad \text{subj. to } \quad & (\theta - \theta^k)^T H(\theta^k) (\theta - \theta^k) \leq \delta,
\end{split}
\end{equation}
where $H$ is the Hessian matrix of the constraint in (\ref{trpo_objec_1}) with respect to $\theta.$ Using duality theory, this problem has the analytical solution
\begin{equation}\label{analytical_trpo_update}
\theta^{k+1} = \theta^{k} + \alpha \sqrt{\frac{2\delta}{g^T H^{-1} g}}H^{-1}g,
\end{equation}
where $\alpha$ is determined through backtracking line search and $g = \nabla_\theta \mathcal{L}(\theta, \theta^k) \Big\rvert_{\theta = \theta^k}.$

Computing or storing $H^{-1}$ may be prohibitively expensive, but according to the form of (\ref{analytical_trpo_update}), finding $H^{-1}g$ is sufficient, and computing or storing this matrix-vector product is relatively cheap. TRPO uses the conjugate gradient method to obtain $H^{-1}g$ while circumventing computation or storage of $H^{-1}$, enabling its use in practical settings.

\subsubsection{Trust Region Policy Optimization as approximated Mirror Descent}

\cite{neu2017unified} shows that the TRPO update with the form
\begin{multline}\label{trpo_objec}
\pi_{k+1} = \argmax_{\pi} \left\{ \sum_{x} \nu_{\pi_k}(s) \sum_{a} \pi(a \mid s) \left( \vphantom{\frac{\pi(a \mid s)}{\pi_k(a \mid s)}} R(s,a) \right. \right. \\ 
\left. \left. + \sum_{s'} P(s' \mid s,a) V^{\pi_k}_{\eta=\infty}(s') - V^{\pi_k}_{\eta=\infty}(s) - \frac{1}{\eta} \log{\frac{\pi(a \mid s)}{\pi_k(a \mid s)}} \right) \right\}
\end{multline}
approximates mirror descent and that this update can be expressed in closed form as
$$\pi_{k+1}(a \mid s) \propto \pi_{k}(a \mid s)e^{\eta \big( R(s,a) + \sum_{s'} P(s' \mid s,a) V^{\pi_k}_{\eta=\infty}(s') - V^{\pi_k}_{\eta=\infty}(s) \big) }.$$
\cite{schulman2015trust} claims that the theoretical TRPO updates are monotonic improvements, but do not claim whether TRPO converges to an optimal policy. However, through duality theory, \cite{neu2017unified} shows that the theoretical TRPO updates indeed converge to the optimal policy $\pi^*$.

\subsection{A3C as Dual Averaging}

\cite{neu2017unified} shows that the A3C algorithm \cite{mnih2016asynchronous} aims to optimize the objective 
$$ \sum_{x} \nu_{\pi_k}(s) \sum_{a} \pi(a \mid s) \left( R(s,a) + \sum_{s'} P(s' \mid s,a) V^{\pi_k}_{\eta=\infty}(s') - V^{\pi_k}_{\eta=\infty}(s) - \frac{1}{\eta^k} \log{\pi_\theta(a \mid s)} \right),$$
which can be interpreted as a dual-averaging \cite{xiao2010dual, duchi2011dual} variant of the TRPO objective in (\ref{trpo_objec}).

Again through the use of duality theory, \cite{neu2017unified} conjectures that A3C does \emph{not} converge and proposes a modified method with convergence guarantees.

\subsection{Constrained Policy Optimization (CPO)}

\cite{achiam2017constrained} uses an approximated update similar to (\ref{nat_grad}) but with an additional constraint that is affine in $\theta$ to make safety guarantees that hold in expectation. The dual problem in CPO is similar to that of TRPO, and the optimal update to $\theta$ can be derived analytically through duality theory here as well.

\subsection{A Lyapunov-based approach to Safe Reinforcement Learning}

\cite{chow2018lyapunov} puts forth a primal-dual subgradient method that obtains a policy that satisfies certain safety constraints in expectation.

\subsection{Guided Policy Search (GPS)}

\cite{levine2013guided} can be used to train complex high-dimensional policies without optimizing the policies directly in high-dimensional parameter spaces. GPS trains a student policy to imitate a teacher policy that is also being trained to produce behaviors that the the student can demonstrate. This approach formulates the Lagrangian of a specified cost function and takes gradient steps on primal and dual variables in an alternating fashion.

Furthermore, \cite{montgomery2016guided} shows that GPS can be seen as an approximate variant of mirror descent. More generally, \cite{peters2010relative, zimin2013online} show that the form of the update in GPS can formulated as mirror descent with the Bregman divergence $D_S$ induced by the relative entropy of the stationary state-action distribution $\mu_\pi$ corresponding to the policy $\pi$.

\section{Conclusion}

The study and application of duality theory has led to key advances in the understanding and efficiency of solving optimization problems in a number of fields. However, duality hasn't received comparable focus in reinforcement learning even though much of the influential work in RL has relied on it. We've shown that duality arises more often in RL than one may think, with benefits including increased interpretability, convergence guarantees, and state-of-the-art advances in algorithmic performance and efficiency. Going forward we hope others will more strongly consider duality while addressing reinforcement learning problems.

\printbibliography

\end{document}